\documentclass[sigconf]{acmart}

\usepackage{booktabs} % For formal tables

\copyrightyear{2018} 
\acmYear{2018} 
\setcopyright{acmcopyright}
\acmConference[MM '18]{2018 ACM Multimedia Conference}{October 22--26, 2018}{Seoul, Republic of Korea}
\acmBooktitle{2018 ACM Multimedia Conference (MM '18), October 22--26, 2018, Seoul, Republic of Korea}
\acmPrice{15.00}
\acmDOI{10.1145/3240508.3240713}
\acmISBN{978-1-4503-5665-7/18/10}

\begin{document}
\title{Deep Adaptive Temporal Pooling for Activity Recognition}

\author{Sibo Song}
\affiliation{%
  \institution{Singapore University of Technology and Design}
  \city{Singapore}
  \country{Singapore}
}
\email{sibo_song@mymail.sutd.edu.sg}

\author{Ngai-Man Cheung}
\affiliation{%
  \institution{Singapore University of Technology and Design}
  \city{Singapore}
  \country{Singapore}
}
\email{ngaiman_cheung@sutd.edu.sg}

\author{Vijay Chandrasekhar}
\affiliation{%
  \institution{Institute for Infocomm Research}
  \city{Singapore}
  \country{Singapore}
}
\email{vijay@i2r.a-star.edu.sg}

\author{Bappaditya Mandal}
\affiliation{%
  \institution{Keele University}
  \city{Keele}
  \state{Staffordshire}
  \country{United Kingdom}
}
\email{b.mandal@keele.ac.uk}

\renewcommand{\shortauthors}{Sibo Song et al.}

\begin{abstract}
Deep neural networks have recently achieved competitive accuracy for human activity recognition.
However, there is room for improvement, especially in modeling of long-term temporal importance and determining the activity relevance of different temporal segments in a video.
To address this problem, we propose a learnable and differentiable module: Deep Adaptive Temporal Pooling (DATP). 
DATP applies a self-attention mechanism to adaptively pool the classification scores of different video segments.  
Specifically, using frame-level features, DATP regresses  importance of different temporal segments, and  generates weights for them. Remarkably, {\em DATP is trained using only the video-level label}. There is no need of  additional supervision except video-level activity class label.
We conduct extensive experiments to investigate various input features and different weight models. 
Experimental results show that DATP can learn to assign large weights to  key video segments. More importantly, DATP can improve training of frame-level feature extractor. This is because  relevant temporal segments are assigned large weights during back-propagation.
Overall, we achieve state-of-the-art performance on UCF101, HMDB51 and Kinetics datasets. 
\end{abstract}


\maketitle

%-------------------------------------------------------------------------

\section{Introduction}
\label{sec:introduction}

% Figure
\begin{figure}
\begin{center}
\includegraphics[width=1.\columnwidth]{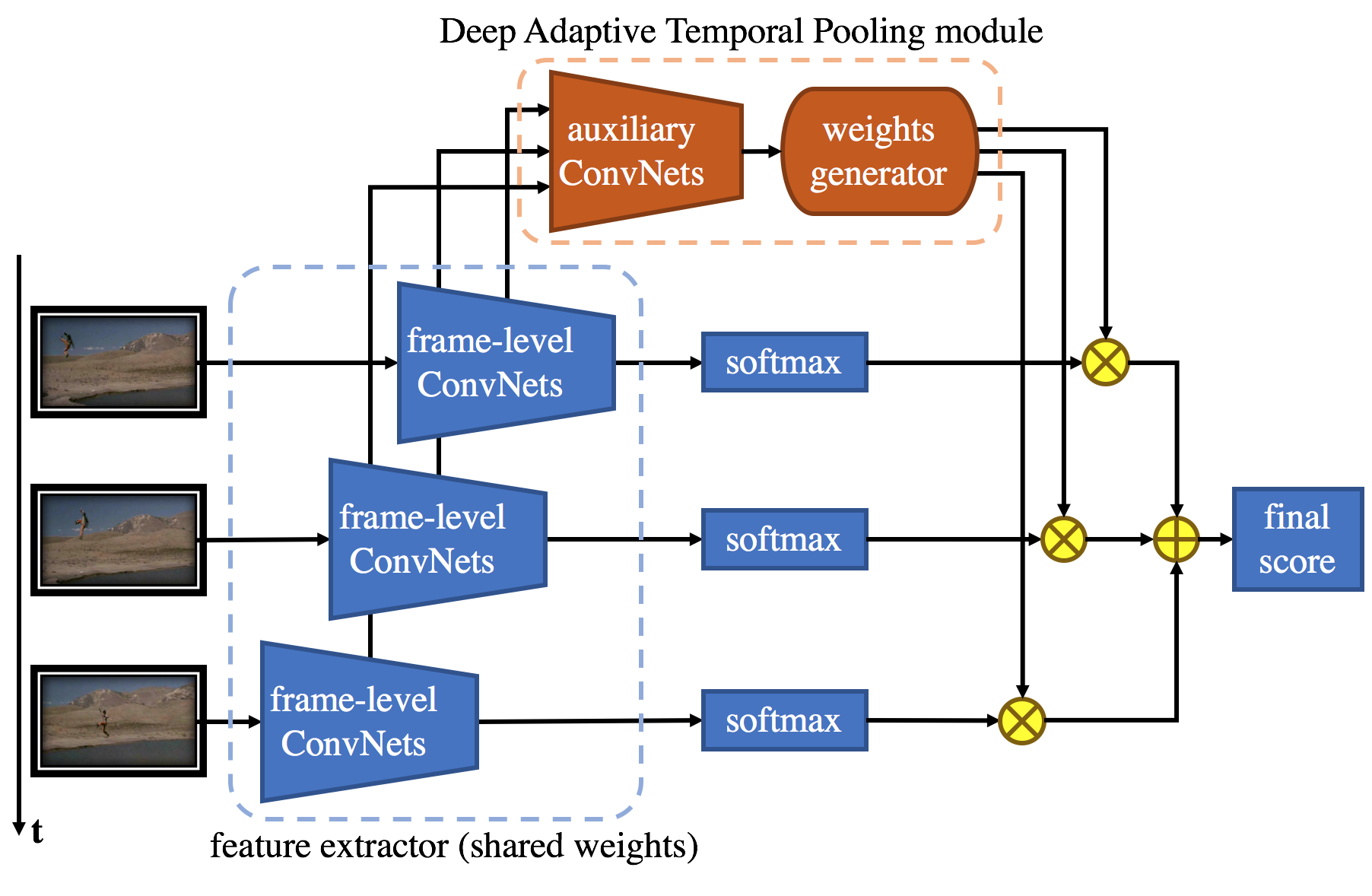}
\caption{\textbf{Pipeline:} A video is divided into $N$ temporal segments. A pair of RGB frame and stacking optical flows is randomly selected from each segment as the input to the frame-level ConvNets. The frame-level Convnets extract intermediate features and compute the softmax scores. The generated softmax scores  are then adaptively pooled by DATP module to obtain the video-level prediction.
\textbf{DATP:} The auxiliary ConvNets consist of convolutional layers and fully-connected layers. DATP takes  intermediate features as inputs, and regresses parameters of a weight model (e.g., Gaussian). Then, the weight generator samples weights from the regressed model. The weights are assigned to different temporal segments. The weighted softmax scores are then summed to obtain the final video-level score.}
\label{fig:diagram_03}
\end{center}
\end{figure}

Recognizing human activities from videos is a key research area in video processing and artificial intelligence. Due to the recent success of deep neural networks (DNN), researchers are using deep learning to solve video processing problems like activity recognition \cite{donahue2015long,simonyan2014two,wang2016actions,zhu2016key}, video captioning \cite{donahue2015long,yao2015describing,venugopalan2015sequence,zhou2017adaptive}, etc.

ConvNets based action recognition can be  categorized into two groups, C3D \cite{tran2015learning} and two-stream ConvNets \cite{simonyan2014two}.
Authors in \cite{tran2015learning} propose 3D ConvNets to learn spatio-temporal features. They utilize 3D volume filters instead of 2D filters. However, it is inadequate in  modeling long-term temporal structures. 
Another successful ConvNets is two-stream ConvNets. It matches the performance of state-of-the-art  hand-crafted features like dense trajectories.  
The two streams are spatial and temporal nets. The spatial stream extracts high-level appearance features. The temporal stream uses optical flow as input to learn motion patterns. 
However, stacking optical flows could only capture short-term temporal cues of actions. Therefore, researchers have proposed several methods to learn spatio-temporal features based on the two-stream architecture \cite{feichtenhofer2016spatiotemporal,wangspatiotemporal,kar2016adascan,diba2017deep,feichtenhofer2016convolutional,feichtenhofer2017spatiotemporal}. 
There are other attempts in modeling temporal structures by incorporating Recurrent Neural Network (RNN) and Long Short-Term Memory (LSTM). \cite{baccouche2011sequential,donahue2015long,wang2016hierarchical,pei2016temporal}

In this paper, we propose a novel module to capture long-term temporal information in ConvNets. 
There are several attempts to design specialized DNN layers for certain tasks.  
In \cite{jaderberg2015spatial}, the author introduces Spatial Transformer Networks (STN) to achieve spatial invariance and identify spatial attention for a given image.
Authors in \cite{girshick2015fast} propose RoI (region of interest) pooling layer to convert the features inside a valid region of interest into a small feature map for localization task. These work motivate us to design DNN layers that can exploit temporal structure and achieve adaptive pooling. 

Specifically, we propose Deep Adaptive Temporal Pooling (DATP) module to model long-term temporal structure as illustrated in Figure \ref{fig:diagram_03}.  
This module is a novel application of STN idea \cite{jaderberg2015spatial} which has been found to be useful in many computer vision tasks, e.g. person re-identification \cite{guo2018efficient}. 
We identify temporal attention and relevant temporal segments in a video for activity recognition with DATP module. 
It consists of auxiliary ConvNets, weight generator and weighting module.
Given input frame-level features, the auxiliary ConvNets regress weights and assign the weights to temporal segments.
In particular, the generated weights can adapt to temporal translation and scaling of key actions. Thus, our system can achieve  temporally invariant in a parameter-efficient manner. Moreover, as will be discussed, DATP module is computationally-efficient, incurs  little computation overhead, and can be implemented easily. Furthermore, DATP is flexible and universal: it can be inserted into different points of a neural network; it can be used with hand-crafted features to determine activity relevance of temporal segments.

%-------------------------------------------------------------------------
\section{Related Work}

\textbf{Traditional machine learning methods} Before deep learning approach emerges, many earlier methods relied on combining hand-crafted features, embedding techniques and Support Vector Machine (SVM) classifier to analyze video data. 
One of the earliest attempts \cite{laptev2008learning} represented videos using Bag-of-Visual-Words which embed HOG features (Histogram of Gradients) and HOF (Histogram of Flow) features with a dictionary. 
There are likewise other spatio-temporal features like SIFT3D \cite{scovanner20073}, MBH \cite{dalal2006human} that are proposed to build better representation for capturing motion and appearance information. 
Recently trajectory-based approach becomes popular for activity recognition. Improved Dense Trajectory \cite{wang2013action} is the state-of-the-art among all hand-crafted features. 
It has shown significant improvement over other existing hand-crafted approaches combining with Fisher kernel framework \cite{sanchez2013image}. 

\textbf{Deep learning approaches} Recently, many deep architectures are proposed to solve video classification problem. 
The very earliest attempt, in 2011, \cite{baccouche2011sequential} combines ConvNets and RNN for human action recognition. 
\cite{karpathy2014large} trains a deep network with video frames from a large dataset for recognizing sports activities. However, analyzing still images only lacks temporal structure and motion information for activity recognition. 
\cite{simonyan2014two} deals with this problem and utilize optical flow to capture short-term motion cues. Specifically, the authors propose to combine spatial and temporal streams which operate on RGB frames and stacking optical flows separately to overcome the lack of temporal information.
In another direction of tackling this problem, \cite{tran2015learning} extends a 2D ConvNets to a 3D ConvNets which utilizes 3D volume filters to enable the networks to learn temporal structures.
\cite{varol2017long} further extends 3D filters with longer temporal dimensions to capture more temporal information.
I3D \cite{carreira2017quo} is recently proposed to efficiently capture spatio-temporal features by inflating the 2D convolutional kernels into 3D kernels.  
\cite{donahue2015long} overcomes the problem that previous work cannot encode long-term temporal information by combining RNN. The authors directly connect ConvNets to RNN model and jointly train them simultaneously to learn temporal dynamics and perceptual representations. 
\cite{wang2016actions} also highlights the sequential information of activity, and designs a Siamese network that models action as a transformation on a high-level feature space.  
Similarly, \cite{kar2016adascan} continuously predicted the discriminative importance of each frame and subsequently pooled them to achieve online classification. 
\cite{diba2017deep} presents Temporal Linear Encoding (TLE) to temporally aggregate features sparsely sampled over the whole video with a bilinear model. 
In \cite{wangspatiotemporal}, authors model correlations between two streams hierarchically by compact bilinear layer at multiple levels.
Recently, \cite{tran2018closer} improves the performance by factorizing 3D convolutional kernels into spatial and temporal filters.
\cite{zhou2018mict} integrates 2D filters with the 3D convolutional filters to learn better spatio-temporal features.
\cite{cherian2018non} captures the action dynamics by utilizing kernelized subspace representations.

\textbf{Temporal attention} Our work has similar objective as  \cite{diba2017deep,wangspatiotemporal,wang2017untrimmednets}  to model long-term temporal dependency using  temporal attention mechanism. However, the approaches are different. 
\cite{diba2017deep} exploits feature interactions between the segments of an entire video. It  linearly encodes and aggregates information. 
\cite{wangspatiotemporal} introduces a compact bilinear operator to temporally fuse multi-path optical flow features. 
\cite{wang2017untrimmednets} learns the weights directly with linear transformation. In contrast, 
we propose a ConvNet module to regress the parameters of weight model.
Recently, \cite{vaswani2017attention} introduces a Transformer module for machine translation. They apply  self-attention mechanism and calculate the dependency of each word by taking whole sequence into consideration. 
Our work is similar to this work in spirit: we compute the temporal weights for video segments without any additional information. 

\textbf{Knowledge distillation} An intuitive way to pool the video segments together is to simply average the probabilities produced by the frame-level classifier\cite{wang2016temporal,varol2017long,qiu2017learning,carreira2017quo}. Knowledge distillation \cite{ba2014deep,hinton2015distilling} is initially proposed to train a network using probability vectors instead of hard labels transferred from another since they usually provide more information per training case and much less variance. Inspired by \cite{hinton2015distilling}, we for the first propose to distill the knowledge temporally by digging into the temporal sequence of probabilities or other high-level intermediate features to model long-term temporal structure for human activity recognition.

% Figure,
\begin{figure*}
\centering
\includegraphics[width=2.03\columnwidth]{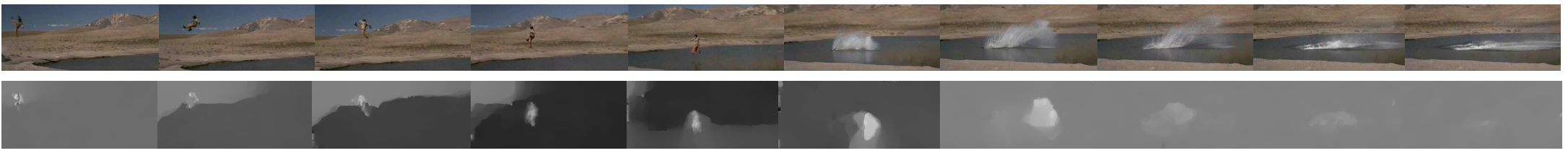}  \\
\includegraphics[width=2.03\columnwidth]{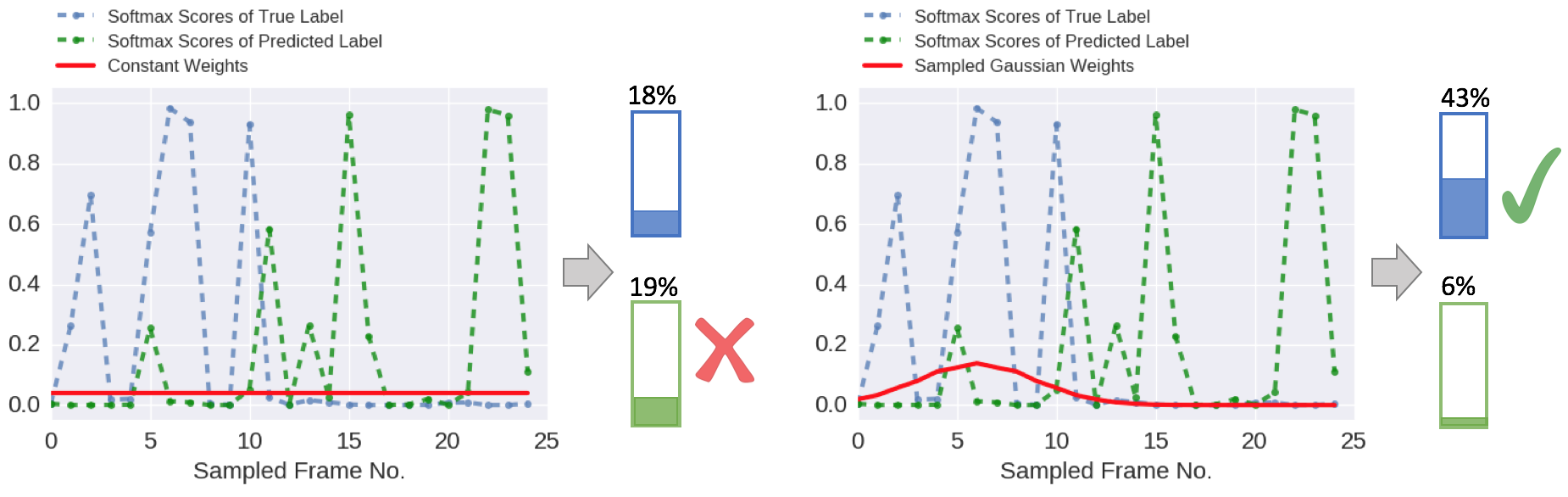}
\caption{\textbf{Top}: A misclassified video sample using TSN: \textit{dive} activity of the HMDB51. We display sampled RGB frames and only horizontal optical flows for simplicity. 
\textbf{Bottom}: Softmax scores of ground-truth activity class \textit{dive} (blue dashed line) and predicted class \textit{fall floor} (green dashed line). Average pooling (left) results in misclassification. In contrast,  adaptive pooling using our DATP (right) assigns large weights to temporal segments in the first half of the video, results in correct classification. 
We emphasize that only video-level class labels are used to train the DATP module.
Best viewed in color.
}
\label{fig:diagram_04}
\end{figure*}

%-------------------------------------------------------------------------
\section{Proposed Approach}

In this section, our motivation of proposing the DATP module is first presented. Then we describe its three components, auxiliary ConvNets, weights generator and weighting strategy with their implementations of forward and backward pass for training.

%-------------------------------------------------------------------------
\subsection{Motivation}

For many existing approaches \cite{wang2016temporal,varol2017long,scovanner20073}, they simply label all frames as same as video-level labels, in other words, viewing these frames with equal contributions to activity recognition. 
However, this will raise a mislabeling issue for irrelevant volumes of the videos if we cannot detect them first, especially for mid-scale datasets like UCF101 and HMDB51 which have limited training data to enable the networks to acquire excellent generalization capability. Key frames detection and classification is a chicken-and-egg problem. To solve this problem, we propose a new learnable and differentiable module which distills activity contributions from semantic features and allows detection by classification in an end-to-end training. Our proposed approach can simultaneously identify some key frames and assign adaptive weights during pooling. 

From frames in Figure \ref{fig:diagram_04}, we can see that the man is falling into a lake. Then we present softmax scores from TSN's temporal stream of the top 2 predicted activities, \textit{dive} (blue dashed line) and \textit{fall floor} (green dashed line). When this man reaches the lake, the optical flow input data produce high softmax scores in \textit{fall floor} activity which makes sense since it causes very similar optical flow changes. Therefore, this sample is misclassified as \textit{fall floor} when using average pooling. While, if we impose proper unimodal Gaussian weights on both softmax scores, the network can successfully classify the sample as \textit{dive} activity (shown in Figure \ref{fig:diagram_04}) on the temporal stream. Our goal is to automatically regress such weights with the DATP module. 

%-------------------------------------------------------------------------
\subsection{Deep Adaptive Temporal Pooling Module}
\label{sec:module}

Given a sequence of intermediate features extracted from frame-level ConvNets as input, our goal is to compute an importance score as a weight for every softmax score of the sampled video segments. To achieve this, we propose the DATP module which consists of three parts (see Figure \ref{fig:diagram_03}): auxiliary ConvNets, weights generator and weighting strategy. 
First, an auxiliary ConvNets takes the intermediate features as input, and through a few stacked layers outputs the parameters of a weights generator. The outputs are then used to parameterize a weights generator model where the temporal weights are sampled from to weight the softmax scores. In the end, temporal weights and softmax scores are combined following the weighting strategy which will be discussed later. 
Note that, although we choose neural networks as the base framework, DATP is a universal pooling module that can exploit temporal structures over any sequential data, e.g., hand-crafted features.

%-------------------------------------------------------------------------
\textbf{Auxiliary ConvNets} The goal of the auxiliary ConvNets is to learn from sequences of high-level features to generate parameters that define the weights generator model. It consists of convolutional layers and fully connected layers with a final regression layer at the end to produce the model parameters of weights generator. 

We first sample $N$ temporal segments from each video sample. Let us consider $F({\boldsymbol{x} }_{ 1,\dots ,N})$ as an underlying mapping that transforms the intermediate features to the parameters of weights generator. It can be approximated by a number of hidden layers with ${ \boldsymbol{x} }_{ 1,\dots ,N }$ denoting the input $\left\{ { \boldsymbol{x} }_{ 1 },\dots ,{ \boldsymbol{x} }_{ N } \right\} $ of length $N$. The input features can take any layer's output which means we can vary where the DATP module is inserted into the overall model. The generated parameters are defined as $\theta =F({\boldsymbol{x} }_{ 1,\dots ,N})$, and their sizes vary depending on different types of the parameterized models.

%-------------------------------------------------------------------------
\textbf{Weights generator} The weights generator model combining with auxiliary ConvNets can be seen as a self-attention module. By taking advantage of high-level features sequences, e.g., probability sequences, it enables the network to distill the contribution score and attend to key actions without additional knowledge. Two implementations of the weights generator model will be discussed, discrete weights model and mixture of Gaussians. 

\begin{itemize}
\item \textbf{Discrete weights model:} To perform a weighting of softmax scores, we need to generate $N$ temporal weights for $N$ video segments. Therefore, one straightforward idea is to simply generate $N$ discrete weights from auxiliary ConvNets and directly assign them to the softmax scores. In this case, the weights generator is an identity function and the weights are the same as regressed parameters, denoted by $\boldsymbol{w} = \left\{ { w }_{ 1 },\dots ,{ w }_{ N } \right\} = \theta \in { \mathbb{R} }^{ N }$.

\item \textbf{Mixture of Gaussians (MoG):} Video segments of activities are usually related to each other and have a certain temporal structure. Therefore, it might be difficult to learn to directly regress discrete weights for the segments. For this purpose, we define a mixture of Gaussians as weights generator model. We take probability density function of the MoG as the form of weights generator. Then we sample weights from it and assign them to different segments. 
\end{itemize}

There are several reasons for choosing the MoG model: 
First, the transitions between non-action and action volumes (i.e., action starting and action ending) are usually smooth, which each Gaussian can model very well. 
Second, the parameters that define Gaussian distribution, mean and variance values, can be perfectly adapted to the temporal translation and scaling (i.e. duration) of key actions. This superb property enables the networks to be temporally invariant to time-series data like videos in a parameter-efficient manner. 
Third, the p.d.f of the MoG is differentiable with respect to its parameters and input, which is critical since it allows gradients to be back-propagated to update the whole model by end-to-end training.

Note that, since we focus on activity classification instead of detection, it is sufficient to recognize and assign larger weights for just several informative temporal segments. 
For example, max pooling practically works well for image classification which simply forwards the highest value of the local patches and only route the gradients to it. 
The same logic applies for activity classification with our DATP module as we only need to forwards the most discriminative information. 
Selecting all informative segments might further improve the accuracy, but selecting some can already achieve significant improvements. 
Furthermore, only highly weighted segments will be updated significantly during training and it allows better training of frame-level ConvNets. 

Formally, given the softmax score $ { \boldsymbol{I} }_{ i } \in {\mathbb{R}}^{C}$ extracted from $N$ video segment at time ${ t }_{ i },\quad i=1,2,\dots ,N$. Note that, we normalize the ${ t }_{ i }$ values into $\left[ 0, 1 \right]$. The number of Gaussians is denoted as $K$. The corresponding temporal weights are 

\small
\begin{equation}
{ w }_{ i }=\sum _{ k }^{ K }{ \frac { { \rho  }_{ k } }{ \sqrt { 2\pi { { \sigma  } }_{ k }^{ 2 } }  } e^{ -\frac { ({ t }_{ i }-{ \mu  }_{ k }) }{ 2{ { \sigma  } }_{ k }^{ 2 } }  },i=1,2,\dots ,N } 
\end{equation}
\normalsize
which is generated from a mixture of $K$ Gaussian distribution of mixture weight $\left\{ { \rho }_{ 1 },\dots ,{ \rho }_{ K } \right\}$, mean $\left\{ { \mu }_{ 1 },\dots ,{ \mu }_{ K } \right\}$ and variance $\left\{ { \sigma }_{ 1 },\dots ,{ \sigma }_{ K } \right\}$. 

To allow backpropagation of gradients on weights generator, we can define the gradients with respect to $\theta \in {\mathbb{R}}^{3\times K} $, \textit{i.e.}, the parameters of mixture of Gaussians, 

\small
\begin{equation}
\begin{aligned}
& \frac { \partial { w }_{ i } }{ \partial { \rho  }_{ j } } ={ \rho  }_{ j }{ \mathcal{N} }_{ { t }_{ i } }({ \mu  }_{ j },{ \sigma  }_{ j }^{ 2 })\frac { 1 }{ { \rho  }_{ j } }, \\
&  \frac { \partial { w }_{ i } }{ \partial { \mu  }_{ j } } ={ { \rho  }_{ j }{ \mathcal{N} }_{ { t }_{ i } }({ \mu  }_{ j },{ \sigma  }_{ j }^{ 2 }) }\frac { ({ t }_{ i }-{ \mu  }_{ j }) }{ { { \sigma  }_{ j } }^{ 2 } }, \\
&  \frac { \partial { w }_{ i } }{ \partial { \sigma  }_{ j } } ={ \rho  }_{ j }{ \mathcal{N} }_{ { t }_{ i } }({ \mu  }_{ j },{ \sigma  }_{ j }^{ 2 })\frac { 1 }{ { { \sigma  }_{ j } } } \left[ \frac { { ({ t }_{ i }-{ \mu  }_{ j }) }^{ 2 } }{ { { \sigma  }_{ j } }^{ 2 } } -1 \right], \\
\end{aligned}
\end{equation}
\normalsize

In the forward pass, given mixture weights, mean value, variance and ${t}_{i}$, the weights generator calculates weights from a mixture of Gaussians. For backpropagation, the generator will update these parameters according to the cross-entropy loss between pooled scores and ground-truth labels. 

%-------------------------------------------------------------------------
\textbf{Weighting strategy} As aforementioned, to explicitly encourage peaky distributions of softmax scores over time, we apply a linear combination between Gaussian weights and temporal softmax scores to produce final pooled representation. 
By training on thousands of activity videos, the learnable module will emphasize the most informative frames by assigning larger temporal weights from the parameterized mixture of Gaussians to prominent temporal segments. 
During training, the mixture weights, mean and variance will be updated to fit the softmax scores of true labels. Therefore, the DATP module can magnify the softmax scores of key segments in the final pooled representation.

Therefore, we define the pooling function as

\small
\begin{equation}
\begin{aligned}
\boldsymbol{S} = \frac { \sum_{i=1}^{N} { \boldsymbol{ I }_{ i } } \times { w }_{ i } }{ {\lVert \boldsymbol{w} \rVert}_{2}  } 
\end{aligned}
\end{equation}
\normalsize

where $\boldsymbol{S} \in {\mathbb{R}}^{C}$ denotes pooled vector. To allow backpropagation of the gradients, we can define the gradients with respect to the input softmax score ${ \boldsymbol{I} }_{ i }$ and the Gaussian weights $ { w }_{ i } $,

\small
\begin{equation}
\begin{aligned}
& \frac { \partial \boldsymbol{ S } }{ \partial \boldsymbol{ I }_{ i } } =\frac { { w }_{ i } }{ {\lVert \boldsymbol{w} \rVert}_{2}  } \mathbf{1} \\
& \frac { \partial \boldsymbol{ S } }{ \partial { w }_{ i } } = \sum_{k = 1}^{C} \frac { {\lVert \boldsymbol{w} \rVert}_{2}^{ 2 } \boldsymbol{ I }_{ i }[k] - { { w }_{ i } } ({ \sum_{j = 1}^{N} { { w }_{ j } \boldsymbol{ I }_{ j}[k] } }) }{ {\lVert \boldsymbol{w} \rVert}_{2}^{ 3 } } 
\end{aligned}
\label{eq:eq_04}
\end{equation}
\normalsize

where $\lVert \boldsymbol{w} \rVert$ denotes the $l_{2}$-norm of the weights vector $\boldsymbol{w}$. 

%-------------------------------------------------------------------------
\textbf{Benefits for training and inference} The advantages of integrating the DATP module are twofold. First, the generated temporal weights are able to highlight key actions and suppress irrelevant frames and therefore boost activity recognition performance. Second, it assists the training of frame-level ConvNets and results in an improved classifier.

The first benefit is intuitive. During inference, the frame-level feature extractor produces a softmax score for each segment.
As is often the case, key actions occur momentarily during the entire video. Then the softmax scores of true label might have high values only when key actions happen. 
Therefore, the video is very likely to be misclassified if we simply average all softmax scores.
However, if informative frames can be identified and their weights are distilled from high-level feature sequences, we will stand a good chance of classifying activities correctly since the frames are weighted adaptively according to their temporal importance.

More importantly, the DATP module can assist training a better frame-level feature extractor. 
If average pooling is applied during training, each segment will contribute equally to update the frame-level ConvNets no matter whether the segment is related to the true activity class or not. 
Different from average pooling, our proposed module can assign weights from adaptive Gaussians and linearly combine them with softmax scores temporally. Therefore, during the backward pass, the frame-level ConvNets are supposed to be updated with different coefficients according to the weights from forward pass (refer to Eq. (\ref{eq:eq_04})). 
Consequently, the frame-level ConvNets has a better discriminating ability and we further confirm that by conducting experiments in Section \ref{sec:results}.

%------------------------------------------------------------------------
\section{Experiments}

In this section, we first describe three challenging datasets, HMDB51, UCF101 and Kinetics. Second, we present implementation details of our evaluation framework. Third, we provide ablation studies to show the benefits of incorporating DATP module and conduct extensive experiments to choose input features from various intermediate layers and investigate possible models as weights generator. Finally, we visualize the temporal weights that network adaptively learns and compare our proposed DATP module with other deep learning methods.

\begin{table}[tbp]
\begin{center}
\scalebox{1.}{
\begin{tabular}{l|c|c|c}
\toprule
input features    & input dim.      & DATP arch.      & Accuracy (\%)   \\ \midrule 
softmax scores    & [10, C]         & 2 conv + 2 fc   & 65.4            \\ 
conv features     & [10, 7, 7, 512] & 3 conv + 2 fc   & \textbf{66.0}   \\
conv features     & [10, 7, 7, 512] & 3 conv + 3 fc   & 65.9            \\
\bottomrule
\end{tabular}}
\end{center}
\caption{Results on the temporal stream of 1\textsuperscript{st} split of HMDB51 using different input features.}
\label{tab:features}
\end{table}

%-------------------------------------------------------------------------
\subsection{Datasets}

UCF101 \cite{soomro2012ucf101} dataset consists of 13,320 video clips from 101 activity categories. All clips have fixed frame rate and resolution of 25 FPS and $320 \times 240$ respectively. 
The length of clips ranges from 1s to 70s. We report classification accuracy according to the experimental setup of 3 train/test splits recommended by \cite{soomro2012ucf101} to keep consistent with other reported results on this dataset. 

HMDB51 \cite{kuehne2011hmdb} contains 6,766 video clips extracted from various sources such as YouTube and movies. It has a total of 51 distinct activity categories, each containing 101 clips at least. 
Many videos selected from videos in this dataset contain scene transitions and severe camera motion which are very challenging for the optical-flow-based approaches. 
We follow the 3 selected splits provided by the authors for evaluation.

Kinetics \cite{kay2017kinetics} is a large-scale video dataset of diverse human activities. It consists of approximately 250k training video clips and 20k validation clips from 400 human action classes which has an order of magnitude more videos than previous datasets. Our model is trained on the whole training set and test on the validation set.

%-------------------------------------------------------------------------
\subsection{Implementation details}

To compare with most of the existing approaches, we choose Temporal Segment Network (TSN) \cite{wang2016temporal} as the baseline approach and experimental framework for two-stream networks. All models are trained on 2 Titan X Pascals GPUs. We employ a pre-trained ResNet34 model \cite{he2016deep} trained on the ImageNet dataset \cite{deng2009imagenet} as backbone model. The two-stream network consists of a spatial and a temporal stream. The spatial ConvNet takes a single RGB frame as the input, and the temporal ConvNet takes 10 stacking optical flow. The dimension of input data is $224 \times 224$ for training. Random cropping and horizontal flipping are employed to augment training data. We choose $N=10$ segments in both training and testing to obtain more temporal information. 
These segments are sampled from videos with a fixed length and random offset. 
Note that TSN chooses 3 as the number of segments during training, while 25 for testing in order to improve performance. Instead, we choose a consistent scheme for both training and testing. We adopt a late fusion strategy for fusing two streams. Specifically, two pooled vectors are generated from each stream, and we take a weighted average of them by setting the spatial weight as 1 and temporal weight as 1.5. 

\begin{table}[tbp]
\begin{center}
\scalebox{1.}{
\begin{tabular}{c|c|c}
\toprule
Input   & softmax scores: [10, C]               & conv features: [10, 7, 7, 512]        \\ 
\midrule
conv1   & [3, 1, 1] conv, 32 ReLU               & [1, 7, 7, 512] conv, 64 ReLU          \\
conv2   & [3, C, 32] conv, 32 ReLU              & [3, 1, 1, 64] conv, 32 ReLU           \\
conv3   & -                                     & [3, 1, 1, 32] conv, 32 ReLU           \\
fc1     & fc layer (after flattening)           & fc layer (after flattening)           \\ 
fc2     & fc layer                              & fc layer                              \\
\bottomrule
\end{tabular}}
\end{center}
\caption{Architectures of the proposed DATP module.}
\label{tab:module}
\end{table}

We use dense optical flow approach for extracting motions from videos as the input of temporal stream. TVL1 algorithm \cite{zach2007duality} is chosen as the implementation of optical flow algorithm. For MoG parameter initialization, we initialize the mean value to [$1/(K+1)$, ... , $K/(K+1)$] and the variance to 0.2 for mixture of $K$ Gaussians. 
For example, the mean value is set to 0.5 for single Gaussian model. 
We implement the proposed DATP module in PyTorch. For experiments on Kinetics dataset, we use per-trained models downloaded from \cite{tsn2016page} and then fine-tuned the model with DATP module inserted.
It can be dropped into a ConvNets architecture at any point. This module is computationally economical and causes very little time overhead when used with high-level features as input. For more details, please refer to Section \ref{sec:ablation} since the architecture varies due to different input and weights generator model. 

%-------------------------------------------------------------------------
\subsection{Ablation studies}
\label{sec:ablation}

In this section, we seek to answer three important questions in utilizing the DATP module. First, we study the effect of different locations to insert the module. Second, we investigate various possible models for generating temporal weights. Finally, we compare different testing schemes with various models.

\textbf{Input features for temporal weight regression} We first study the effect of different input features of DATP module on the activity recognition performance. In other words, we vary where the DATP module is inserted into the overall model to distill temporal information. Here, we explore two choices of intermediate features for feeding DATP module, (1) \textbf{softmax scores:} output from softmax function after last fully connected layer and (2) \textbf{conv features:} output from last convolutional layers just before global average pooling layer. Therefore, conv features keep the spatial dimension of $7 \times 7$ while softmax scores do not. Note that, we only investigate the high-level features of the network since low-level features generally need much deeper auxiliary ConvNets to generate reasonable weights which is expensive comparing high-level features such as softmax scores. The auxiliary ConvNets comprises two temporal convolutional layers for both softmax scores and conv features as the input features. For conv features, we add one extra convolutional layer of size $7 \times 7$ before temporal convolution to reduce the dimensionality. Moreover, we vary the number of fully connected layers to investigate the effect of the networks' depth. 

Table \ref{tab:features} and \ref{tab:module} provide details about the input dimensionality and architectures of DATP module and reports the performance on the temporal stream of the 1\textsuperscript{st} split of HMDB51 as we vary the input features. Note that, we use a unimodal Gaussian, \textit{i.e.}, MoG of $K=1$ for all results in this table. It clearly shows the benefit of replacing softmax scores by conv features as the input to DATP module. Furthermore, increasing number of fully connected layers does not exhibit any superior performance. Therefore, we choose conv features for the following experiments as our evaluation baseline.

\textbf{Model for weights generator} We evaluate two aforementioned models for generating temporal weights, 
(1) \textbf{Discrete weights model:} We modify the number of output of auxiliary ConvNets to $N$. Then, we directly take these $N$ discrete weights as the temporal weights of $N$ softmax scores in the overall model. Finally, the pooled score of each stream is a linear combination of $N$ weights and $N$ softmax scores.
(2) \textbf{Mixture of Gaussians (MoG):} Unimodal Gaussian is an effective and reasonable model which can be viewed as a soft version of max pooling since it temporal-invariantly addresses key actions during training and inference. Therefore, we expect MoG will further improve the accuracy since MoG could model more complex temporal structures and unimodal Gaussian is a special case of the MoG. Specifically, we explore the impact of different number of Gaussians on the classification performance.

\begin{table}[t]
\begin{center}
\scalebox{1.0}{
\begin{tabular}{l|c}
\toprule
model                       &   Accuracy (\%)   \\ \midrule
discrete weights            &   63.1            \\ 
Mixture of Gaussians (K=1)  &   66.0            \\ 
Mixture of Gaussians (K=2)  &   \textbf{66.4}   \\ 
Mixture of Gaussians (K=3)  &   65.6            \\ 
\bottomrule
\end{tabular}}
\end{center}
\caption{Comparing different models of the weights generator on the temporal stream of 1\textsuperscript{st} split of HMDB51.}
\label{tab:weights_generator}
\end{table}

In Table \ref{tab:weights_generator}, we report the accuracy on the temporal stream of the 1\textsuperscript{st} split of HMDB51 for different models of the weights generator. For MoG, we choose $K=1, 2, 3$ as the number of Gaussians. By comparing the performance, we observe that using a mixture of 2 Gaussians achieves the highest accuracy which produces a slightly better result than unimodal Gaussian. However, the model degrades classification performance when $K=3$ which implies that simply increasing the number of Gaussians might impair the performance. Moreover, all MoG models constantly outperform the model that generates discrete weights. We conjecture that the decrease in accuracy is due to more parameters in discrete weights model and the mixture of 3 Gaussians. Although both models should be able to asymptotically regress the desired weights, the ease of learning might be different.

%-------------------------------------------------------------------------
\textbf{Temporally trained frame-level ConvNets} In \cite{wang2016temporal}, authors choose 3 segments during training, while 25 for testing in order to improve test performance. In our framework, we keep the DATP module when testing and a consistent number of segments for both training and test unlike \cite{wang2016temporal}. 
While, since the frame-level ConvNets shares parameters in our framework, the trained model can be viewed as a frame-level feature extractor and performs frame-wise evaluation without appending DATP module during testing (\cite{simonyan2014two,wang2016temporal}). Therefore, we compare different pooling strategies to show the effectiveness of DATP module and its advantages over average pooling. In Table \ref{tab:testing}, we summarized the performance on 1\textsuperscript{st} train/test split of HMDB51 with the following testing schemes:

% Table
\begin{table}[t]
\begin{center}
\scalebox{1.0}{
\begin{tabular}{l|c|c|c}
\toprule
testing scheme & Spatial   & Temporal  & Fusion  \\ \midrule
P1             & 48.9      & 54.1      & 58.3    \\ 
P2             & 56.8      & 66.0      & 72.3    \\ 
P3             & 54.4      & 62.4      & 69.5    \\ 
DATP           & 57.1      & 66.4      & 72.9    \\ 
\bottomrule
\end{tabular}}
\end{center}
\caption{Results on 1\textsuperscript{st} split of HMDB51 of different testing schemes.}
\label{tab:testing}
\end{table}

% Figure
\begin{figure*}
\centering
\includegraphics[width=2.03\columnwidth]{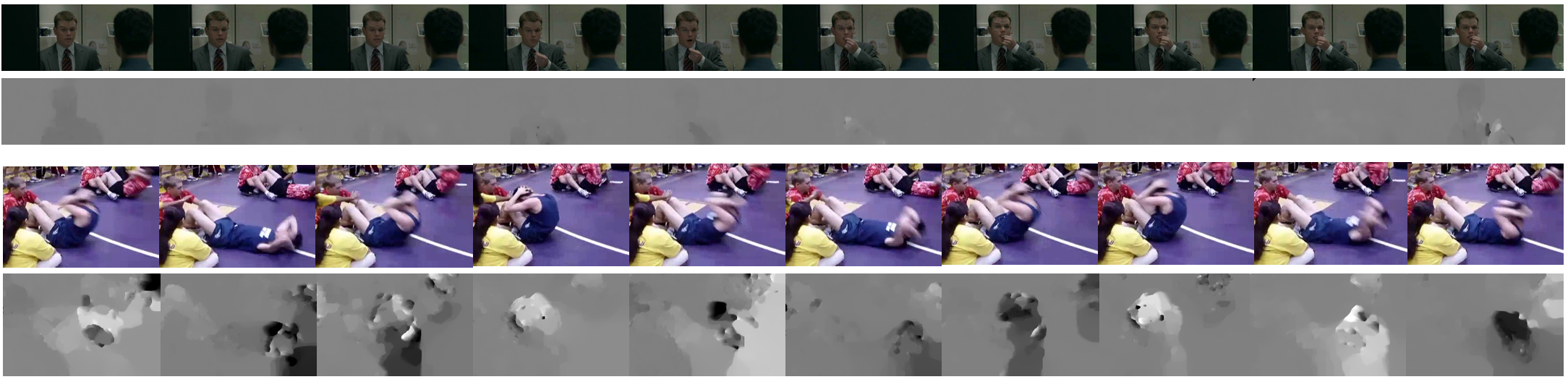} \\
\includegraphics[width=2.03\columnwidth]{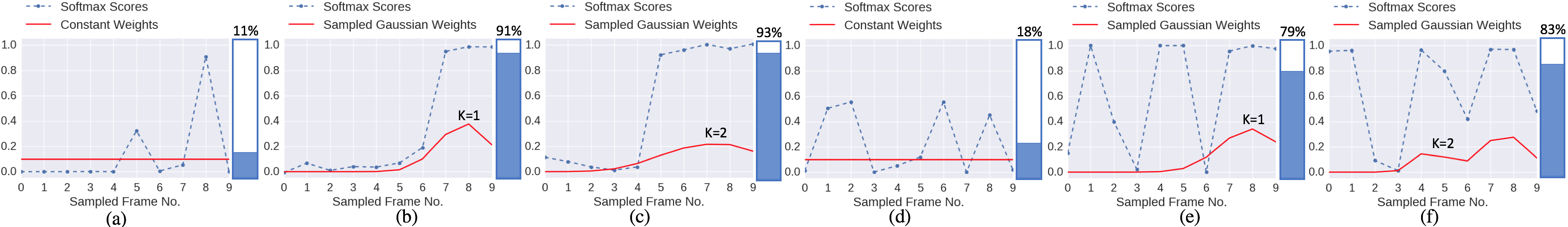}
\caption{\textbf{Top:} Visualization of video samples of activity \textit{eat} and \textit{sit-up}. RGB frames and horizontal optical flows of 10 sampled temporal segments are presented. \textbf{Bottom: } Softmax scores and sampled weights generated from DATP module: (a) \textit{eat} (baseline), (b) \textit{eat} (DATP with 1 Gaussian), (c) \textit{eat} (DATP with 2 Gaussians), (d) \textit{sit-up} (baseline), (e) \textit{sit-up} (DATP with 1 Gaussian), (f) \textit{sit-up} (DATP with 2 Gaussians).
Note that the  softmax scores {\em before} applying the weights are shown (blue dashed lines).
The improved softmax scores in (b), (c), (e), (f) demonstrate the improvement in the temporally trained ConvNets using DATP. These improved softmax scores are then adaptively pooled using the generated weights (red lines) to obtain the final video-level score.
}
\label{fig:diagram_06}
\end{figure*}

\begin{itemize}
\item \textbf{P1 (temporally trained ConvNets + random Gaussians): } We train the model with DATP module which takes conv features as input and mixture of 2 Gaussians as weights generator model. Then we employ the improved frame-level ConvNets as the feature generator. While during testing, we generate a mixture of 2 random Gaussians instead of taking adaptive Gaussian weights directly from DATP.
\item \textbf{P2 (temporally trained ConvNets + average pooling): } We train the whole model with DATP module as same as P1. Then 25 RGB frames and optical flow stacks are sampled from the video following the test setup of \cite{wang2016temporal} and \cite{simonyan2014two}. Finally, we average the 25 softmax scores as the final prediction.
\item \textbf{P3 (naively trained ConvNets + average pooling): } We train the model without DATP module and take the trained feature extractor with average pooling for testing. This is similar to the TSN framework but with more segments for training.
\item \textbf{DATP (temporally trained ConvNets + adaptive Gaussians): } The proposed DATP framework is used for both training and testing.  
\end{itemize}

By comparing P2 with P3, it is clear that the frame-level ConvNet has been improved for classification using DATP module since P2 and P3 both apply average pooling during the testing stage. However, P2 utilizes a frame-level ConvNet trained with DATP while P3 does not. 
Moreover, we observe an improvement from P2 to DATP where the improved ConvNets are used for both, while P2 applies average pooling and DATP applies adaptive temporal pooling for testing. 
It is worth noting that P2 samples 25 segments for testing whereas DATP takes 10. Additionally, we evaluate another setting P1 that uses a mixture of 2 random Gaussians (random value of mean and variance) as weights generator during testing. 
This result confirms that our proposed auxiliary ConvNets can regress meaningful model parameters from the input intermediate features. 

%-------------------------------------------------------------------------
\subsection{Results and Analysis}
\label{sec:results}

In this section, we first present visualizations of the temporal weights from different pooling techniques. Second, we report improvements in each category by using DATP. Finally, we compare our proposed approach against state of the art. All experiments are done with the conv features as input and the mixture of 2 Gaussians model as weights generator. 

%-------------------------------------------------------------------------
\textbf{Visualization of generated temporal weights} Generating explicit temporal weights is one of our key contributions which enables our proposed module to work as a weak action detector. Therefore, we visualize the intermediate results - generated temporal weights in two typical cases presented in Figure \ref{fig:diagram_06} of activity \textit{eat} and \textit{sit-up}.
We show the softmax scores and generated weights from the baseline model (average pooling), DATP with MoG (K=1) and DATP with MoG (K=2). They are only extracted from the temporal stream for easy comparison.

From the \textit{eat} activity sample, we can see that two men are talking to each other, while one of them is eating during the last five sampled frames. Figure \ref{fig:diagram_06} (b) and (c) clearly shows that the trained frame-level ConvNets produces high responses for the last four to five frames. Then the DATP module generates unimodal and mixture of Gaussians weights respectively which place the peak values around the 8th frame and therefore increases the probability of \textit{eat} activity by pooling all temporal softmax scores. 

From a gradient perspective, we can see that the gradients can be much higher with the action-relevant segments during the backward pass since the weights simply represent the coefficients associated with the gradients. Therefore, it allows our approach to exhibit a higher ability to discriminate key actions than average pooling. This has been proved in Figure \ref{fig:diagram_06} (a) to (c), since we observe significant improvements of the feature extractors. In addition, we can clearly see the benefit of training with a mixture of 2 Gaussians. Not only it generates an improved feature extractor, but also it can better weights the flat region of last five frames' softmax scores. 

Similarly, in the second video of activity \textit{sit-up}, the \textit{sit-up} action arises periodically and both unimodal Gaussian and mixture of 2 Gaussians successfully recognize the key actions. Furthermore, a mixture of 2 Gaussians model can capture more information from complex temporal structures. We notice that even with more than 2 peaks in the softmax scores, we can still classify the activity correctly, as some peak has been assigned a larger temporal weight. Improvement can be achieved without selecting all informative video segments as we discussed in Section \ref{sec:module}.

%-------------------------------------------------------------------------
\textbf{Impact on different classes} We compare the changes of classification accuracy by applying DATP module over the baseline method on the 1\textsuperscript{st} train/test split of HMDB51 using the temporal stream to show the effect of DATP module for each action category. 

We find that the largest increase occur in \textit{kick}, \textit{push} and \textit{shoot\_gun} categories which is $26.7\%$. Besides, \textit{smile} and \textit{fall\_floor} have around $20\%$ improvement. Note that the numbers here refer to the absolute changes of correctly predicted ratio per category.
For better performance on atomic actions such as \textit{kick}, \textit{push} and \textit{smile}, we believe the reason is that these actions usually last a short period of time and DATP is able to capture and emphasize the occurrence. 
Similarly, for action \textit{shoot gun}, changes in optical flow occur within a short time and in a small part of the scene as well which is challenging for activity recognition with average pooling technique. However, our method can highlight key frames to predict the label even when they are far less compared to uninformative frames. 
On the other hand, for actions such as \textit{talk} and \textit{jump}, DATP does not show advantages over the baseline approach. We conjecture this is due to the fact that these activities are continuously evolving and many video samples have a complicated background for recognition, e.g., jumping on staircases might be misclassified as \textit{climb stairs}.

%-------------------------------------------------------------------------
\textbf{Computational overhead} We show in our approach description that DATP does not require any modification of the baseline network and incurs very little computational cost when attached to existing architectures. In a basic setting with single Gaussian as weights generator and softmax scores as input, the running time is $\sim 0.75 s/batch$ without DATP and $\sim 0.83 s/batch$ with DATP(Implemented with PyTorch on 2 Titan Xp GPU). This proves that the DATP module can improve the training, increase classification accuracy at a small additional computational cost less than 10\%. It is much preferred when compared with spatio-temporal convolutions since 3D convolutions are generally inefficient.

% Table
\begin{table}[t]
\begin{center}
\scalebox{1.0}{
\begin{tabular}{l|c|c}
\toprule
 Accuracy (\%)                                      & UCF101  & HMDB51 \\  \midrule
Two-stream \cite{simonyan2014two}                   & 88.8    & 59.4   \\ 
C3D \cite{tran2015learning}                         & 82.3    & 56.8   \\ 
Long-term Temporal Convolution \cite{varol2017long} & 91.7    & 64.8   \\ 
KVMF \cite{zhu2016key}                              & 93.1    & 63.3   \\ 
Transformations \cite{wang2016actions}              & 92.4    & 63.4   \\ 
ConvFusion \cite{feichtenhofer2016convolutional}    & 92.5    & 65.4   \\
ST-ResNet \cite{feichtenhofer2016spatiotemporal}    & 93.4    & 66.4   \\ 
I3D \cite{carreira2017quo}                          & 93.4    & 66.4   \\
ActionVLAD \cite{girdhar2017actionvlad}             & 92.7    & 66.9   \\
AdaScan \cite{kar2016adascan}                       & 93.2    & 66.9   \\ 
TSN \cite{wang2016temporal}                         & 94.0    & 68.5   \\ 
ST-Multiplier \cite{feichtenhofer2017spatiotemporal}& 94.2    & 68.9   \\
ST-Pyramid  \cite{wangspatiotemporal}               & 94.6    & 68.9   \\
ST-VLMPF(DF) \cite{dutaspatio}                      & 93.6    & 69.5   \\
TLE:Bilinear \cite{diba2017deep}                    & 95.6    & 71.1   \\ \midrule
DATP (1 Gaussian + softmax scores)                  & 95.1    & 71.6   \\
DATP (2 Gaussians + conv features)                  & \textbf{95.9}  & \textbf{72.3}  \\ 
\bottomrule
\end{tabular}}
\end{center}
\caption{Comparison with existing deep learning methods on UCF101 and HMDB51 dataset. For fair comparison, we report results from methods that do not pre-train on the Kinetics dataset.}
\label{tab:table_03}
\end{table}

%-------------------------------------------------------------------------
\textbf{Comparison against state of the art} We compare our approach with existing deep learning methods in Table \ref{tab:table_03} and Table \ref{tab:table_04} for UCF101, HMDB51 and Kinetics datasets. We can see that DATP achieves state-of-the-art results on all these three datasets with conv features and mixture model of 2 Gaussians. It is noteworthy that, by inserting the lightweight DATP module into TSN model, our proposed architecture outperforms the original model by a good margin on all three datasets. The improved performance demonstrates superiority of DATP and its effectiveness on temporal knowledge distillation. 

Note that, AdaScan \cite{kar2016adascan} and TLE \cite{diba2017deep} also address long-term temporal information for activity recognition. 
Notably, TLE exploits long-term dynamics by capturing interactions between the segments and encodes them linearly into a compact representation for video-level prediction, while our approach makes use of the non-linear mixture of Gaussian model which is temporally invariant to actions. They model action transitions for different activities, while we model the temporal importance which can directly boost the performance. 
AdaScan adaptively pools and represents frames in an online fashion. It predicts an importance score of each frame to determine their contributions to the final pooled descriptor. However, the difference is that AdaScan predicts the importance score at each time only with previously pooled features. While our DATP extracts information across the whole video to decide the temporal weights. In addition, our DATP is a universal pooling module that can exploit temporal structures over any sequential data, not limited to deep features. Therefore, we can also incorporate it with hand-craft features, such as dense trajectories \cite{wang2013dense} or TDD \cite{wang2015action} features. 

% Table
\begin{table}[t]
\begin{center}
\scalebox{1.}{
\begin{tabular}{l|c|c}
\toprule
 Accuracy (\%)                                      & Top-1   & Top-5  \\ \midrule
ARTNet \cite{wang2017appearance} spatial stream     & 70.7    & 89.3   \\ 
ARTNet \cite{wang2017appearance} temporal stream    & 60.6    & 83.1   \\ 
ARTNet \cite{wang2017appearance} fusion             & 72.4    & 90.4   \\ \midrule
I3D \cite{carreira2017quo} spatial stream           & 71.1    & 89.3   \\  
I3D \cite{carreira2017quo} temporal stream          & 63.4    & 84.9   \\  
I3D \cite{carreira2017quo} fusion                   & 74.2    & 91.3   \\ \midrule
TSN \cite{wang2016temporal} spatial stream          & 72.5    & 90.2   \\ 
TSN \cite{wang2016temporal} temporal stream         & 62.8    & 84.2   \\ 
TSN \cite{wang2016temporal} fusion                  & 76.6    & 92.4   \\ \midrule
DATP spatial stream                                 & 72.9    & 91.6   \\ 
DATP temporal stream                                & 63.9    & 85.6   \\ 
DATP fusion                                         & \textbf{77.1}  & \textbf{93.1}  \\ 
\bottomrule
\end{tabular}}
\end{center}
\caption{Comparison with existing deep learning methods on Kinetics dataset. We use mixture of 2 Gaussians and conv features as input to DATP for Kinetics evaluation. (Note that, the performance of TSN above is reported in \cite{tsn2016page})}
\label{tab:table_04}
\end{table}

%------------------------------------------------------------------------
\section{Conclusions}

In this paper, we propose a Deep Adaptive Temporal Pooling module (DATP) to capture long-term temporal information. Our DATP module allows for self-attention and temporal knowledge distillation. It utilizes a MoG model to compute adaptive weights to pool temporal segments together without extra supervision. DATP incurs little computational overhead and can be easily implemented. We investigated various input features for temporal weight regression and several weights generator models.
We showed that the DATP module contributes to training of an improved feature extractor. Our work achieves state-of-the-art performance. 

%-------------------------------------------------------------------------

\bibliographystyle{ACM-Reference-Format}
\bibliography{sample-bibliography}


\begin{thebibliography}{49}

%%% the Web address.

\ifx \showCODEN    \undefined \def \showCODEN     #1{\unskip}     \fi
\ifx \showDOI      \undefined \def \showDOI       #1{#1}\fi
\ifx \showISBNx    \undefined \def \showISBNx     #1{\unskip}     \fi
\ifx \showISBNxiii \undefined \def \showISBNxiii  #1{\unskip}     \fi
\ifx \showISSN     \undefined \def \showISSN      #1{\unskip}     \fi
\ifx \showLCCN     \undefined \def \showLCCN      #1{\unskip}     \fi
\ifx \shownote     \undefined \def \shownote      #1{#1}          \fi
\ifx \showarticletitle \undefined \def \showarticletitle #1{#1}   \fi
\ifx \showURL      \undefined \def \showURL       {\relax}        \fi
% invisible to TeX
\providecommand\bibfield[2]{#2}
\providecommand\bibinfo[2]{#2}
\providecommand\natexlab[1]{#1}
\providecommand\showeprint[2][]{arXiv:#2}

\bibitem[\protect\citeauthoryear{Ba and Caruana}{Ba and Caruana}{2014}]%
        {ba2014deep}
\bibfield{author}{\bibinfo{person}{Jimmy Ba} {and} \bibinfo{person}{Rich
  Caruana}.} \bibinfo{year}{2014}\natexlab{}.
\newblock \showarticletitle{Do deep nets really need to be deep?}. In
  \bibinfo{booktitle}{\emph{Advances in neural information processing
  systems}}. \bibinfo{pages}{2654--2662}.
\newblock


\bibitem[\protect\citeauthoryear{Baccouche, Mamalet, Wolf, Garcia, and
  Baskurt}{Baccouche et~al\mbox{.}}{2011}]%
        {baccouche2011sequential}
\bibfield{author}{\bibinfo{person}{Moez Baccouche}, \bibinfo{person}{Franck
  Mamalet}, \bibinfo{person}{Christian Wolf}, \bibinfo{person}{Christophe
  Garcia}, {and} \bibinfo{person}{Atilla Baskurt}.}
  \bibinfo{year}{2011}\natexlab{}.
\newblock \showarticletitle{Sequential deep learning for human action
  recognition}. In \bibinfo{booktitle}{\emph{International Workshop on Human
  Behavior Understanding}}. Springer, \bibinfo{pages}{29--39}.
\newblock


\bibitem[\protect\citeauthoryear{Carreira and Zisserman}{Carreira and
  Zisserman}{2017}]%
        {carreira2017quo}
\bibfield{author}{\bibinfo{person}{Joao Carreira} {and} \bibinfo{person}{Andrew
  Zisserman}.} \bibinfo{year}{2017}\natexlab{}.
\newblock \showarticletitle{Quo vadis, action recognition? a new model and the
  kinetics dataset}. In \bibinfo{booktitle}{\emph{2017 IEEE Conference on
  Computer Vision and Pattern Recognition (CVPR)}}. IEEE,
  \bibinfo{pages}{4724--4733}.
\newblock


\bibitem[\protect\citeauthoryear{Cherian, Sra, Gould, and Hartley}{Cherian
  et~al\mbox{.}}{2018}]%
        {cherian2018non}
\bibfield{author}{\bibinfo{person}{Anoop Cherian}, \bibinfo{person}{Suvrit
  Sra}, \bibinfo{person}{Stephen Gould}, {and} \bibinfo{person}{Richard
  Hartley}.} \bibinfo{year}{2018}\natexlab{}.
\newblock \showarticletitle{Non-Linear Temporal Subspace Representations for
  Activity Recognition}. In \bibinfo{booktitle}{\emph{Proceedings of the IEEE
  Conference on Computer Vision and Pattern Recognition}}.
  \bibinfo{pages}{2197--2206}.
\newblock


\bibitem[\protect\citeauthoryear{Dalal, Triggs, and Schmid}{Dalal
  et~al\mbox{.}}{2006}]%
        {dalal2006human}
\bibfield{author}{\bibinfo{person}{Navneet Dalal}, \bibinfo{person}{Bill
  Triggs}, {and} \bibinfo{person}{Cordelia Schmid}.}
  \bibinfo{year}{2006}\natexlab{}.
\newblock \showarticletitle{Human detection using oriented histograms of flow
  and appearance}. In \bibinfo{booktitle}{\emph{European conference on computer
  vision}}. Springer, \bibinfo{pages}{428--441}.
\newblock


\bibitem[\protect\citeauthoryear{Deng, Dong, Socher, Li, Li, and Fei-Fei}{Deng
  et~al\mbox{.}}{2009}]%
        {deng2009imagenet}
\bibfield{author}{\bibinfo{person}{Jia Deng}, \bibinfo{person}{Wei Dong},
  \bibinfo{person}{Richard Socher}, \bibinfo{person}{Li-Jia Li},
  \bibinfo{person}{Kai Li}, {and} \bibinfo{person}{Li Fei-Fei}.}
  \bibinfo{year}{2009}\natexlab{}.
\newblock \showarticletitle{Imagenet: A large-scale hierarchical image
  database}. In \bibinfo{booktitle}{\emph{Computer Vision and Pattern
  Recognition, 2009. CVPR 2009. IEEE Conference on}}. IEEE,
  \bibinfo{pages}{248--255}.
\newblock


\bibitem[\protect\citeauthoryear{Diba, Sharma, and Van~Gool}{Diba
  et~al\mbox{.}}{2017}]%
        {diba2017deep}
\bibfield{author}{\bibinfo{person}{Ali Diba}, \bibinfo{person}{Vivek Sharma},
  {and} \bibinfo{person}{Luc Van~Gool}.} \bibinfo{year}{2017}\natexlab{}.
\newblock \showarticletitle{Deep temporal linear encoding networks}. In
  \bibinfo{booktitle}{\emph{Computer Vision and Pattern Recognition}}.
\newblock


\bibitem[\protect\citeauthoryear{Donahue, Anne~Hendricks, Guadarrama, Rohrbach,
  Venugopalan, Saenko, and Darrell}{Donahue et~al\mbox{.}}{2015}]%
        {donahue2015long}
\bibfield{author}{\bibinfo{person}{Jeffrey Donahue}, \bibinfo{person}{Lisa
  Anne~Hendricks}, \bibinfo{person}{Sergio Guadarrama}, \bibinfo{person}{Marcus
  Rohrbach}, \bibinfo{person}{Subhashini Venugopalan}, \bibinfo{person}{Kate
  Saenko}, {and} \bibinfo{person}{Trevor Darrell}.}
  \bibinfo{year}{2015}\natexlab{}.
\newblock \showarticletitle{Long-term recurrent convolutional networks for
  visual recognition and description}. In \bibinfo{booktitle}{\emph{Proceedings
  of the IEEE conference on computer vision and pattern recognition}}.
  \bibinfo{pages}{2625--2634}.
\newblock


\bibitem[\protect\citeauthoryear{Duta, Ionescu, Aizawa, and Sebe}{Duta
  et~al\mbox{.}}{2017}]%
        {dutaspatio}
\bibfield{author}{\bibinfo{person}{Ionut~Cosmin Duta}, \bibinfo{person}{Bogdan
  Ionescu}, \bibinfo{person}{Kiyoharu Aizawa}, {and} \bibinfo{person}{Nicu
  Sebe}.} \bibinfo{year}{2017}\natexlab{}.
\newblock \showarticletitle{Spatio-Temporal Vector of Locally Max Pooled
  Features for Action Recognition in Videos}. In
  \bibinfo{booktitle}{\emph{Proceedings of the IEEE conference on Computer
  Vision and Pattern Recognition}}.
\newblock


\bibitem[\protect\citeauthoryear{Feichtenhofer, Pinz, and Wildes}{Feichtenhofer
  et~al\mbox{.}}{2016a}]%
        {feichtenhofer2016spatiotemporal}
\bibfield{author}{\bibinfo{person}{Christoph Feichtenhofer},
  \bibinfo{person}{Axel Pinz}, {and} \bibinfo{person}{Richard Wildes}.}
  \bibinfo{year}{2016}\natexlab{a}.
\newblock \showarticletitle{Spatiotemporal Residual Networks for Video Action
  Recognition}. In \bibinfo{booktitle}{\emph{Advances in Neural Information
  Processing Systems}}. \bibinfo{pages}{3468--3476}.
\newblock


\bibitem[\protect\citeauthoryear{Feichtenhofer, Pinz, and Wildes}{Feichtenhofer
  et~al\mbox{.}}{2017}]%
        {feichtenhofer2017spatiotemporal}
\bibfield{author}{\bibinfo{person}{Christoph Feichtenhofer},
  \bibinfo{person}{Axel Pinz}, {and} \bibinfo{person}{Richard~P Wildes}.}
  \bibinfo{year}{2017}\natexlab{}.
\newblock \showarticletitle{Spatiotemporal multiplier networks for video action
  recognition}. In \bibinfo{booktitle}{\emph{Proceedings of the IEEE Conference
  on Computer Vision and Pattern Recognition}}.
\newblock


\bibitem[\protect\citeauthoryear{Feichtenhofer, Pinz, and
  Zisserman}{Feichtenhofer et~al\mbox{.}}{2016b}]%
        {feichtenhofer2016convolutional}
\bibfield{author}{\bibinfo{person}{Christoph Feichtenhofer},
  \bibinfo{person}{Axel Pinz}, {and} \bibinfo{person}{Andrew Zisserman}.}
  \bibinfo{year}{2016}\natexlab{b}.
\newblock \showarticletitle{Convolutional two-stream network fusion for video
  action recognition}. In \bibinfo{booktitle}{\emph{Proceedings of the IEEE
  Conference on Computer Vision and Pattern Recognition}}.
  \bibinfo{pages}{1933--1941}.
\newblock


\bibitem[\protect\citeauthoryear{Girdhar, Ramanan, Gupta, Sivic, and
  Russell}{Girdhar et~al\mbox{.}}{2017}]%
        {girdhar2017actionvlad}
\bibfield{author}{\bibinfo{person}{Rohit Girdhar}, \bibinfo{person}{Deva
  Ramanan}, \bibinfo{person}{Abhinav Gupta}, \bibinfo{person}{Josef Sivic},
  {and} \bibinfo{person}{Bryan Russell}.} \bibinfo{year}{2017}\natexlab{}.
\newblock \showarticletitle{ActionVLAD: Learning spatio-temporal aggregation
  for action classification}.
\newblock \bibinfo{journal}{\emph{arXiv preprint arXiv:1704.02895}}
  (\bibinfo{year}{2017}).
\newblock


\bibitem[\protect\citeauthoryear{Girshick}{Girshick}{2015}]%
        {girshick2015fast}
\bibfield{author}{\bibinfo{person}{Ross Girshick}.}
  \bibinfo{year}{2015}\natexlab{}.
\newblock \showarticletitle{Fast r-cnn}. In
  \bibinfo{booktitle}{\emph{Proceedings of the IEEE International Conference on
  Computer Vision}}. \bibinfo{pages}{1440--1448}.
\newblock


\bibitem[\protect\citeauthoryear{Guo and Cheung}{Guo and Cheung}{2018}]%
        {guo2018efficient}
\bibfield{author}{\bibinfo{person}{Yiluan Guo} {and} \bibinfo{person}{Ngai-Man
  Cheung}.} \bibinfo{year}{2018}\natexlab{}.
\newblock \showarticletitle{Efficient and Deep Person Re-Identification Using
  Multi-Level Similarity}. In \bibinfo{booktitle}{\emph{Proceedings of the IEEE
  Conference on Computer Vision and Pattern Recognition}}.
  \bibinfo{pages}{2335--2344}.
\newblock


\bibitem[\protect\citeauthoryear{He, Zhang, Ren, and Sun}{He
  et~al\mbox{.}}{2016}]%
        {he2016deep}
\bibfield{author}{\bibinfo{person}{Kaiming He}, \bibinfo{person}{Xiangyu
  Zhang}, \bibinfo{person}{Shaoqing Ren}, {and} \bibinfo{person}{Jian Sun}.}
  \bibinfo{year}{2016}\natexlab{}.
\newblock \showarticletitle{Deep residual learning for image recognition}. In
  \bibinfo{booktitle}{\emph{Proceedings of the IEEE conference on computer
  vision and pattern recognition}}. \bibinfo{pages}{770--778}.
\newblock


\bibitem[\protect\citeauthoryear{Hinton, Vinyals, and Dean}{Hinton
  et~al\mbox{.}}{2015}]%
        {hinton2015distilling}
\bibfield{author}{\bibinfo{person}{Geoffrey Hinton}, \bibinfo{person}{Oriol
  Vinyals}, {and} \bibinfo{person}{Jeff Dean}.}
  \bibinfo{year}{2015}\natexlab{}.
\newblock \showarticletitle{Distilling the knowledge in a neural network}.
\newblock \bibinfo{journal}{\emph{arXiv preprint arXiv:1503.02531}}
  (\bibinfo{year}{2015}).
\newblock


\bibitem[\protect\citeauthoryear{Jaderberg, Simonyan, Zisserman,
  et~al\mbox{.}}{Jaderberg et~al\mbox{.}}{2015}]%
        {jaderberg2015spatial}
\bibfield{author}{\bibinfo{person}{Max Jaderberg}, \bibinfo{person}{Karen
  Simonyan}, \bibinfo{person}{Andrew Zisserman}, {et~al\mbox{.}}}
  \bibinfo{year}{2015}\natexlab{}.
\newblock \showarticletitle{Spatial transformer networks}. In
  \bibinfo{booktitle}{\emph{Advances in Neural Information Processing
  Systems}}. \bibinfo{pages}{2017--2025}.
\newblock


\bibitem[\protect\citeauthoryear{Kar, Rai, Sikka, and Sharma}{Kar
  et~al\mbox{.}}{2016}]%
        {kar2016adascan}
\bibfield{author}{\bibinfo{person}{Amlan Kar}, \bibinfo{person}{Nishant Rai},
  \bibinfo{person}{Karan Sikka}, {and} \bibinfo{person}{Gaurav Sharma}.}
  \bibinfo{year}{2016}\natexlab{}.
\newblock \showarticletitle{AdaScan: Adaptive Scan Pooling in Deep
  Convolutional Neural Networks for Human Action Recognition in Videos}.
\newblock \bibinfo{journal}{\emph{arXiv preprint arXiv:1611.08240}}
  (\bibinfo{year}{2016}).
\newblock


\bibitem[\protect\citeauthoryear{Karpathy, Toderici, Shetty, Leung, Sukthankar,
  and Fei-Fei}{Karpathy et~al\mbox{.}}{2014}]%
        {karpathy2014large}
\bibfield{author}{\bibinfo{person}{Andrej Karpathy}, \bibinfo{person}{George
  Toderici}, \bibinfo{person}{Sanketh Shetty}, \bibinfo{person}{Thomas Leung},
  \bibinfo{person}{Rahul Sukthankar}, {and} \bibinfo{person}{Li Fei-Fei}.}
  \bibinfo{year}{2014}\natexlab{}.
\newblock \showarticletitle{Large-scale video classification with convolutional
  neural networks}. In \bibinfo{booktitle}{\emph{Proceedings of the IEEE
  conference on Computer Vision and Pattern Recognition}}.
  \bibinfo{pages}{1725--1732}.
\newblock


\bibitem[\protect\citeauthoryear{Kay, Carreira, Simonyan, Zhang, Hillier,
  Vijayanarasimhan, Viola, Green, Back, Natsev, et~al\mbox{.}}{Kay
  et~al\mbox{.}}{2017}]%
        {kay2017kinetics}
\bibfield{author}{\bibinfo{person}{Will Kay}, \bibinfo{person}{Joao Carreira},
  \bibinfo{person}{Karen Simonyan}, \bibinfo{person}{Brian Zhang},
  \bibinfo{person}{Chloe Hillier}, \bibinfo{person}{Sudheendra
  Vijayanarasimhan}, \bibinfo{person}{Fabio Viola}, \bibinfo{person}{Tim
  Green}, \bibinfo{person}{Trevor Back}, \bibinfo{person}{Paul Natsev},
  {et~al\mbox{.}}} \bibinfo{year}{2017}\natexlab{}.
\newblock \showarticletitle{The kinetics human action video dataset}.
\newblock \bibinfo{journal}{\emph{arXiv preprint arXiv:1705.06950}}
  (\bibinfo{year}{2017}).
\newblock


\bibitem[\protect\citeauthoryear{Kuehne, Jhuang, Garrote, Poggio, and
  Serre}{Kuehne et~al\mbox{.}}{2011}]%
        {kuehne2011hmdb}
\bibfield{author}{\bibinfo{person}{Hildegard Kuehne}, \bibinfo{person}{Hueihan
  Jhuang}, \bibinfo{person}{Est{\'\i}baliz Garrote}, \bibinfo{person}{Tomaso
  Poggio}, {and} \bibinfo{person}{Thomas Serre}.}
  \bibinfo{year}{2011}\natexlab{}.
\newblock \showarticletitle{HMDB: a large video database for human motion
  recognition}. In \bibinfo{booktitle}{\emph{Computer Vision (ICCV), 2011 IEEE
  International Conference on}}. IEEE, \bibinfo{pages}{2556--2563}.
\newblock


\bibitem[\protect\citeauthoryear{Laptev, Marszalek, Schmid, and
  Rozenfeld}{Laptev et~al\mbox{.}}{2008}]%
        {laptev2008learning}
\bibfield{author}{\bibinfo{person}{Ivan Laptev}, \bibinfo{person}{Marcin
  Marszalek}, \bibinfo{person}{Cordelia Schmid}, {and}
  \bibinfo{person}{Benjamin Rozenfeld}.} \bibinfo{year}{2008}\natexlab{}.
\newblock \showarticletitle{Learning realistic human actions from movies}. In
  \bibinfo{booktitle}{\emph{Computer Vision and Pattern Recognition, 2008. CVPR
  2008. IEEE Conference on}}. IEEE, \bibinfo{pages}{1--8}.
\newblock


\bibitem[\protect\citeauthoryear{Multimedia-Laboratory-CUHK}{Multimedia-Laboratory-CUHK}{2016}]%
        {tsn2016page}
\bibfield{author}{\bibinfo{person}{Multimedia-Laboratory-CUHK}.}
  \bibinfo{year}{2016}\natexlab{}.
\newblock \bibinfo{title}{TSN Pretrained Models on Kinetics Dataset}.
\newblock
  \bibinfo{howpublished}{\url{http://yjxiong.me/others/kinetics_action/}}.
  (\bibinfo{year}{2016}).
\newblock


\bibitem[\protect\citeauthoryear{Pei, Baltru{\v{s}}aitis, Tax, and Morency}{Pei
  et~al\mbox{.}}{2017}]%
        {pei2016temporal}
\bibfield{author}{\bibinfo{person}{Wenjie Pei}, \bibinfo{person}{Tadas
  Baltru{\v{s}}aitis}, \bibinfo{person}{David~MJ Tax}, {and}
  \bibinfo{person}{Louis-Philippe Morency}.} \bibinfo{year}{2017}\natexlab{}.
\newblock \showarticletitle{Temporal Attention-Gated Model for Robust Sequence
  Classification}.
\newblock  (\bibinfo{year}{2017}).
\newblock


\bibitem[\protect\citeauthoryear{Qiu, Yao, and Mei}{Qiu et~al\mbox{.}}{2017}]%
        {qiu2017learning}
\bibfield{author}{\bibinfo{person}{Zhaofan Qiu}, \bibinfo{person}{Ting Yao},
  {and} \bibinfo{person}{Tao Mei}.} \bibinfo{year}{2017}\natexlab{}.
\newblock \showarticletitle{Learning spatio-temporal representation with
  pseudo-3d residual networks}. In \bibinfo{booktitle}{\emph{2017 IEEE
  International Conference on Computer Vision (ICCV)}}. IEEE,
  \bibinfo{pages}{5534--5542}.
\newblock


\bibitem[\protect\citeauthoryear{S{\'a}nchez, Perronnin, Mensink, and
  Verbeek}{S{\'a}nchez et~al\mbox{.}}{2013}]%
        {sanchez2013image}
\bibfield{author}{\bibinfo{person}{Jorge S{\'a}nchez}, \bibinfo{person}{Florent
  Perronnin}, \bibinfo{person}{Thomas Mensink}, {and} \bibinfo{person}{Jakob
  Verbeek}.} \bibinfo{year}{2013}\natexlab{}.
\newblock \showarticletitle{Image classification with the fisher vector: Theory
  and practice}.
\newblock \bibinfo{journal}{\emph{International journal of computer vision}}
  \bibinfo{volume}{105}, \bibinfo{number}{3} (\bibinfo{year}{2013}),
  \bibinfo{pages}{222--245}.
\newblock


\bibitem[\protect\citeauthoryear{Scovanner, Ali, and Shah}{Scovanner
  et~al\mbox{.}}{2007}]%
        {scovanner20073}
\bibfield{author}{\bibinfo{person}{Paul Scovanner}, \bibinfo{person}{Saad Ali},
  {and} \bibinfo{person}{Mubarak Shah}.} \bibinfo{year}{2007}\natexlab{}.
\newblock \showarticletitle{A 3-dimensional sift descriptor and its application
  to action recognition}. In \bibinfo{booktitle}{\emph{Proceedings of the 15th
  ACM international conference on Multimedia}}. ACM, \bibinfo{pages}{357--360}.
\newblock


\bibitem[\protect\citeauthoryear{Simonyan and Zisserman}{Simonyan and
  Zisserman}{2014}]%
        {simonyan2014two}
\bibfield{author}{\bibinfo{person}{Karen Simonyan} {and}
  \bibinfo{person}{Andrew Zisserman}.} \bibinfo{year}{2014}\natexlab{}.
\newblock \showarticletitle{Two-stream convolutional networks for action
  recognition in videos}. In \bibinfo{booktitle}{\emph{Advances in neural
  information processing systems}}. \bibinfo{pages}{568--576}.
\newblock


\bibitem[\protect\citeauthoryear{Soomro, Zamir, and Shah}{Soomro
  et~al\mbox{.}}{2012}]%
        {soomro2012ucf101}
\bibfield{author}{\bibinfo{person}{Khurram Soomro},
  \bibinfo{person}{Amir~Roshan Zamir}, {and} \bibinfo{person}{Mubarak Shah}.}
  \bibinfo{year}{2012}\natexlab{}.
\newblock \showarticletitle{UCF101: A dataset of 101 human actions classes from
  videos in the wild}.
\newblock \bibinfo{journal}{\emph{arXiv preprint arXiv:1212.0402}}
  (\bibinfo{year}{2012}).
\newblock


\bibitem[\protect\citeauthoryear{Tran, Bourdev, Fergus, Torresani, and
  Paluri}{Tran et~al\mbox{.}}{2015}]%
        {tran2015learning}
\bibfield{author}{\bibinfo{person}{Du Tran}, \bibinfo{person}{Lubomir Bourdev},
  \bibinfo{person}{Rob Fergus}, \bibinfo{person}{Lorenzo Torresani}, {and}
  \bibinfo{person}{Manohar Paluri}.} \bibinfo{year}{2015}\natexlab{}.
\newblock \showarticletitle{Learning spatiotemporal features with 3d
  convolutional networks}. In \bibinfo{booktitle}{\emph{Proceedings of the IEEE
  International Conference on Computer Vision}}. \bibinfo{pages}{4489--4497}.
\newblock


\bibitem[\protect\citeauthoryear{Tran, Wang, Torresani, Ray, LeCun, and
  Paluri}{Tran et~al\mbox{.}}{2018}]%
        {tran2018closer}
\bibfield{author}{\bibinfo{person}{Du Tran}, \bibinfo{person}{Heng Wang},
  \bibinfo{person}{Lorenzo Torresani}, \bibinfo{person}{Jamie Ray},
  \bibinfo{person}{Yann LeCun}, {and} \bibinfo{person}{Manohar Paluri}.}
  \bibinfo{year}{2018}\natexlab{}.
\newblock \showarticletitle{A Closer Look at Spatiotemporal Convolutions for
  Action Recognition}. In \bibinfo{booktitle}{\emph{Proceedings of the IEEE
  Conference on Computer Vision and Pattern Recognition}}.
  \bibinfo{pages}{6450--6459}.
\newblock


\bibitem[\protect\citeauthoryear{Varol, Laptev, and Schmid}{Varol
  et~al\mbox{.}}{2017}]%
        {varol2017long}
\bibfield{author}{\bibinfo{person}{Gul Varol}, \bibinfo{person}{Ivan Laptev},
  {and} \bibinfo{person}{Cordelia Schmid}.} \bibinfo{year}{2017}\natexlab{}.
\newblock \showarticletitle{Long-term temporal convolutions for action
  recognition}.
\newblock \bibinfo{journal}{\emph{IEEE Transactions on Pattern Analysis and
  Machine Intelligence}} (\bibinfo{year}{2017}).
\newblock


\bibitem[\protect\citeauthoryear{Vaswani, Shazeer, Parmar, Uszkoreit, Jones,
  Gomez, Kaiser, and Polosukhin}{Vaswani et~al\mbox{.}}{2017}]%
        {vaswani2017attention}
\bibfield{author}{\bibinfo{person}{Ashish Vaswani}, \bibinfo{person}{Noam
  Shazeer}, \bibinfo{person}{Niki Parmar}, \bibinfo{person}{Jakob Uszkoreit},
  \bibinfo{person}{Llion Jones}, \bibinfo{person}{Aidan~N Gomez},
  \bibinfo{person}{{\L}ukasz Kaiser}, {and} \bibinfo{person}{Illia
  Polosukhin}.} \bibinfo{year}{2017}\natexlab{}.
\newblock \showarticletitle{Attention is all you need}. In
  \bibinfo{booktitle}{\emph{Advances in Neural Information Processing
  Systems}}. \bibinfo{pages}{6000--6010}.
\newblock


\bibitem[\protect\citeauthoryear{Venugopalan, Rohrbach, Donahue, Mooney,
  Darrell, and Saenko}{Venugopalan et~al\mbox{.}}{2015}]%
        {venugopalan2015sequence}
\bibfield{author}{\bibinfo{person}{Subhashini Venugopalan},
  \bibinfo{person}{Marcus Rohrbach}, \bibinfo{person}{Jeffrey Donahue},
  \bibinfo{person}{Raymond Mooney}, \bibinfo{person}{Trevor Darrell}, {and}
  \bibinfo{person}{Kate Saenko}.} \bibinfo{year}{2015}\natexlab{}.
\newblock \showarticletitle{Sequence to sequence-video to text}. In
  \bibinfo{booktitle}{\emph{Proceedings of the IEEE International Conference on
  Computer Vision}}. \bibinfo{pages}{4534--4542}.
\newblock


\bibitem[\protect\citeauthoryear{Wang, Kl{\"a}ser, Schmid, and Liu}{Wang
  et~al\mbox{.}}{2013}]%
        {wang2013dense}
\bibfield{author}{\bibinfo{person}{Heng Wang}, \bibinfo{person}{Alexander
  Kl{\"a}ser}, \bibinfo{person}{Cordelia Schmid}, {and}
  \bibinfo{person}{Cheng-Lin Liu}.} \bibinfo{year}{2013}\natexlab{}.
\newblock \showarticletitle{Dense trajectories and motion boundary descriptors
  for action recognition}.
\newblock \bibinfo{journal}{\emph{International journal of computer vision}}
  \bibinfo{volume}{103}, \bibinfo{number}{1} (\bibinfo{year}{2013}),
  \bibinfo{pages}{60--79}.
\newblock


\bibitem[\protect\citeauthoryear{Wang and Schmid}{Wang and Schmid}{2013}]%
        {wang2013action}
\bibfield{author}{\bibinfo{person}{Heng Wang} {and} \bibinfo{person}{Cordelia
  Schmid}.} \bibinfo{year}{2013}\natexlab{}.
\newblock \showarticletitle{Action recognition with improved trajectories}. In
  \bibinfo{booktitle}{\emph{Proceedings of the IEEE International Conference on
  Computer Vision}}. \bibinfo{pages}{3551--3558}.
\newblock


\bibitem[\protect\citeauthoryear{Wang, Li, Li, and Van~Gool}{Wang
  et~al\mbox{.}}{2017a}]%
        {wang2017appearance}
\bibfield{author}{\bibinfo{person}{Limin Wang}, \bibinfo{person}{Wei Li},
  \bibinfo{person}{Wen Li}, {and} \bibinfo{person}{Luc Van~Gool}.}
  \bibinfo{year}{2017}\natexlab{a}.
\newblock \showarticletitle{Appearance-and-Relation Networks for Video
  Classification}.
\newblock \bibinfo{journal}{\emph{arXiv preprint arXiv:1711.09125}}
  (\bibinfo{year}{2017}).
\newblock


\bibitem[\protect\citeauthoryear{Wang, Qiao, and Tang}{Wang
  et~al\mbox{.}}{2015}]%
        {wang2015action}
\bibfield{author}{\bibinfo{person}{Limin Wang}, \bibinfo{person}{Yu Qiao},
  {and} \bibinfo{person}{Xiaoou Tang}.} \bibinfo{year}{2015}\natexlab{}.
\newblock \showarticletitle{Action recognition with trajectory-pooled
  deep-convolutional descriptors}. In \bibinfo{booktitle}{\emph{Proceedings of
  the IEEE Conference on Computer Vision and Pattern Recognition}}.
  \bibinfo{pages}{4305--4314}.
\newblock


\bibitem[\protect\citeauthoryear{Wang, Xiong, Lin, and Van~Gool}{Wang
  et~al\mbox{.}}{2017c}]%
        {wang2017untrimmednets}
\bibfield{author}{\bibinfo{person}{Limin Wang}, \bibinfo{person}{Yuanjun
  Xiong}, \bibinfo{person}{Dahua Lin}, {and} \bibinfo{person}{Luc Van~Gool}.}
  \bibinfo{year}{2017}\natexlab{c}.
\newblock \showarticletitle{Untrimmednets for weakly supervised action
  recognition and detection}. In \bibinfo{booktitle}{\emph{Proc. CVPR}}.
\newblock


\bibitem[\protect\citeauthoryear{Wang, Xiong, Wang, Qiao, Lin, Tang, and
  Van~Gool}{Wang et~al\mbox{.}}{2016c}]%
        {wang2016temporal}
\bibfield{author}{\bibinfo{person}{Limin Wang}, \bibinfo{person}{Yuanjun
  Xiong}, \bibinfo{person}{Zhe Wang}, \bibinfo{person}{Yu Qiao},
  \bibinfo{person}{Dahua Lin}, \bibinfo{person}{Xiaoou Tang}, {and}
  \bibinfo{person}{Luc Van~Gool}.} \bibinfo{year}{2016}\natexlab{c}.
\newblock \showarticletitle{Temporal segment networks: towards good practices
  for deep action recognition}. In \bibinfo{booktitle}{\emph{European
  Conference on Computer Vision}}. Springer, \bibinfo{pages}{20--36}.
\newblock


\bibitem[\protect\citeauthoryear{Wang, Farhadi, and Gupta}{Wang
  et~al\mbox{.}}{2016a}]%
        {wang2016actions}
\bibfield{author}{\bibinfo{person}{Xiaolong Wang}, \bibinfo{person}{Ali
  Farhadi}, {and} \bibinfo{person}{Abhinav Gupta}.}
  \bibinfo{year}{2016}\natexlab{a}.
\newblock \showarticletitle{Actions\~{} transformations}. In
  \bibinfo{booktitle}{\emph{Proceedings of the IEEE Conference on Computer
  Vision and Pattern Recognition}}. \bibinfo{pages}{2658--2667}.
\newblock


\bibitem[\protect\citeauthoryear{Wang, Long, Wang, and Yu}{Wang
  et~al\mbox{.}}{2017b}]%
        {wangspatiotemporal}
\bibfield{author}{\bibinfo{person}{Yunbo Wang}, \bibinfo{person}{Mingsheng
  Long}, \bibinfo{person}{Jianmin Wang}, {and} \bibinfo{person}{Philip~S Yu}.}
  \bibinfo{year}{2017}\natexlab{b}.
\newblock \showarticletitle{Spatiotemporal Pyramid Network for Video Action
  Recognition}. In \bibinfo{booktitle}{\emph{Proceedings of the IEEE conference
  on Computer Vision and Pattern Recognition}}.
\newblock


\bibitem[\protect\citeauthoryear{Wang, Wang, Tang, O'Hare, Chang, and Li}{Wang
  et~al\mbox{.}}{2016b}]%
        {wang2016hierarchical}
\bibfield{author}{\bibinfo{person}{Yilin Wang}, \bibinfo{person}{Suhang Wang},
  \bibinfo{person}{Jiliang Tang}, \bibinfo{person}{Neil O'Hare},
  \bibinfo{person}{Yi Chang}, {and} \bibinfo{person}{Baoxin Li}.}
  \bibinfo{year}{2016}\natexlab{b}.
\newblock \showarticletitle{Hierarchical attention network for action
  recognition in videos}.
\newblock \bibinfo{journal}{\emph{arXiv preprint arXiv:1607.06416}}
  (\bibinfo{year}{2016}).
\newblock


\bibitem[\protect\citeauthoryear{Yao, Torabi, Cho, Ballas, Pal, Larochelle, and
  Courville}{Yao et~al\mbox{.}}{2015}]%
        {yao2015describing}
\bibfield{author}{\bibinfo{person}{Li Yao}, \bibinfo{person}{Atousa Torabi},
  \bibinfo{person}{Kyunghyun Cho}, \bibinfo{person}{Nicolas Ballas},
  \bibinfo{person}{Christopher Pal}, \bibinfo{person}{Hugo Larochelle}, {and}
  \bibinfo{person}{Aaron Courville}.} \bibinfo{year}{2015}\natexlab{}.
\newblock \showarticletitle{Describing videos by exploiting temporal
  structure}. In \bibinfo{booktitle}{\emph{Proceedings of the IEEE
  international conference on computer vision}}. \bibinfo{pages}{4507--4515}.
\newblock


\bibitem[\protect\citeauthoryear{Zach, Pock, and Bischof}{Zach
  et~al\mbox{.}}{2007}]%
        {zach2007duality}
\bibfield{author}{\bibinfo{person}{Christopher Zach}, \bibinfo{person}{Thomas
  Pock}, {and} \bibinfo{person}{Horst Bischof}.}
  \bibinfo{year}{2007}\natexlab{}.
\newblock \showarticletitle{A duality based approach for realtime TV-L 1
  optical flow}. In \bibinfo{booktitle}{\emph{Joint Pattern Recognition
  Symposium}}. Springer, \bibinfo{pages}{214--223}.
\newblock


\bibitem[\protect\citeauthoryear{Zhou, Sun, Liu, Zha, and Zeng}{Zhou
  et~al\mbox{.}}{2017}]%
        {zhou2017adaptive}
\bibfield{author}{\bibinfo{person}{Yizhou Zhou}, \bibinfo{person}{Xiaoyan Sun},
  \bibinfo{person}{Dong Liu}, \bibinfo{person}{Zhengjun Zha}, {and}
  \bibinfo{person}{Wenjun Zeng}.} \bibinfo{year}{2017}\natexlab{}.
\newblock \showarticletitle{Adaptive Pooling in Multi-Instance Learning for Web
  Video Annotation}. In \bibinfo{booktitle}{\emph{Proceedings of the IEEE
  Conference on Computer Vision and Pattern Recognition}}.
  \bibinfo{pages}{318--327}.
\newblock


\bibitem[\protect\citeauthoryear{Zhou, Sun, Zha, and Zeng}{Zhou
  et~al\mbox{.}}{2018}]%
        {zhou2018mict}
\bibfield{author}{\bibinfo{person}{Yizhou Zhou}, \bibinfo{person}{Xiaoyan Sun},
  \bibinfo{person}{Zheng-Jun Zha}, {and} \bibinfo{person}{Wenjun Zeng}.}
  \bibinfo{year}{2018}\natexlab{}.
\newblock \showarticletitle{MiCT: Mixed 3D/2D Convolutional Tube for Human
  Action Recognition}. In \bibinfo{booktitle}{\emph{Proceedings of the IEEE
  Conference on Computer Vision and Pattern Recognition}}.
  \bibinfo{pages}{449--458}.
\newblock


\bibitem[\protect\citeauthoryear{Zhu, Hu, Sun, Cao, and Qiao}{Zhu
  et~al\mbox{.}}{2016}]%
        {zhu2016key}
\bibfield{author}{\bibinfo{person}{Wangjiang Zhu}, \bibinfo{person}{Jie Hu},
  \bibinfo{person}{Gang Sun}, \bibinfo{person}{Xudong Cao}, {and}
  \bibinfo{person}{Yu Qiao}.} \bibinfo{year}{2016}\natexlab{}.
\newblock \showarticletitle{A key volume mining deep framework for action
  recognition}. In \bibinfo{booktitle}{\emph{Proceedings of the IEEE Conference
  on Computer Vision and Pattern Recognition}}. \bibinfo{pages}{1991--1999}.
\newblock


\end{thebibliography}

\end{document}